\documentclass[12pt]{article}
\usepackage[a4paper,top=1in,bottom=1in, total={6in, 8in}]{geometry}

\usepackage{float}
\usepackage{subcaption}
\usepackage{amsmath}
\usepackage{multirow}
\usepackage{graphicx}
\usepackage{booktabs}
\usepackage{placeins}
\usepackage{wrapfig}

\usepackage{url}

\title{\textbf{Efficient Data-Dependent Learnability}}
\author{
    Yaniv Fogel\\
    School of Electrical Engineering\\
    Tel-Aviv University\\
    \texttt{Yanivfogel@mail.tau.ac.il}
  \and
    Tal Shapira\\
    School of Electrical Engineering\\
    Tel-Aviv University\\
    \texttt{talshapira1@mail.tau.ac.il}
       \and
   Meir Feder \\
   School of Electrical Engineering\\
   Tel-Aviv University\\
   \texttt{meir@eng.tau.ac.il} \\
}

\date{}

\begin{document}

\maketitle

\begin{abstract}
The predictive normalized maximum likelihood (pNML) approach has recently been proposed as the min-max optimal solution to the batch learning problem where both the training set and the test data feature are individuals, known sequences. This approach has yields a learnability measure that can also be interpreted as a stability measure. This measure has shown some potential in detecting out-of-distribution examples, yet it has considerable computational costs. In this project, we propose and analyze an approximation of the pNML, which is based on influence functions. Combining both theoretical analysis and experiments, we show that when applied to neural networks, this approximation can detect out-of-distribution examples effectively. We also compare its performance to that achieved by conducting a single gradient step for each possible label.   
\end{abstract}

\section{Introduction}
\label{sec:Influence_functions}

The recent years have seen a major leap in the use of learning algorithms to solve various problems. In particular, neural networks have achieved state of the art performances in various fields. Despite the empirical success, it seems that classical theoretical tools used to evaluate the generalization capabilities of learning algorithms, such as the VC-dimension and Rademacher complexity, see \cite{zhang2016understanding}. The gap in the theoretical understanding of the performance of neural networks naturally attracts researchers to try and come up with other theoretical tools to evaluate generalization in learning problems.

Recently, an individual approach to the definition of the learning problem has recently been suggested in \cite{fogel2019individual}. In the setting presented in this paper, the learner tries to come up with a probability assignment $q(\cdot|x, z^N)$ that will minimize the following min-max regret:

\begin{equation}
\label{pNML}
\min_{q} \max_{y} \log\left(\frac{p_{\hat{\theta}(z^N,x,y)}(y|x)}{q(y|x;z^{N})}\right)
\end{equation}

where $\hat{\theta}(y,x,z^N) = \arg \max_{\theta} p_\theta(y|x) \prod_{t=1}^N p_{\theta}(y_t|x_t)$, e.g. the learner tries to compete with a genie who knows both the training set, the test data feature and its true outcome, yet does not know which of the examples is the true test. In this case, the min-max optimal solution is:

\begin{equation}
\label{regret}
    q(y|x; z^{N}) = \frac{p_{\hat{\theta}(z^N,x,y)}(y|x)}{\sum_{y'}p_{\hat{\theta}(z^N,x,y')}(y'|x)}
\end{equation}

The derived min-max optimal regret is independent of the true $y$:

\begin{equation}
    R^*(z^N, x) = \log (\sum_{y'}p_{\hat{\theta}(z^N,x,y')}(y'|x)) 
\end{equation}

This solution can be interpreted as a rather intuitive learning algorithm: Given some training set $z^N$ and a new test sample $x$, assume that the true test label is some $y$, compute the maximum likelihood over the whole sequence, and take the probability it gives $y$. Then, repeat this process for each possible $y$ and normalize to get a valid probability assignment.

One may note that the min-max optimal regret \ref{regret} measures, in some sense, the possible change in the prediction due to adding another example. Thus, it bears some resemblance to stability measures such as those discussed at    

The implementation of this approach to the hypothesis class of conventional neural networks has been studied in \cite{bibas2019deep}. In this paper the pNML was implemented by first training a model over the training set, then performing several more iterations when the training also contains the test set for each possible label. Those additional iterations were only used to update the last layer. This implementation has shown some potential in detecting out-of-distribution and even adversarial examples, yet this came with considerable computational costs.

One possible variation of this implementation is to perform a single gradient descent example, and to take into consideration only the test set when computing it. This way the additional computations will only include evaluating the gradient over the specific test sample, updating the weights with it and computing the prediction of the updated network over the test sample.   

Another possible solution to the large amount of additional computations can perhaps be obtained using influence functions, \cite{cook1980influence}, a classic statistical technique that measures the effect of adding an infinitesimally small weight to one of the training points on the loss over another point. This method, which has also been used on classes of deep neural networks in \cite{koh2017influence}, can also be used to evaluate the effect of adding a new training point with infinitesimally small weight. It can also be interpreted as conducting one newton step for this new point. as As we will show, this technique can also be used to get an approximation to the pNML where the weight given to the new example is some small $epsilon$ instead of $1$.

Our goal in this paper is to find an efficient approximation of the pNML using influence function and to evaluate the performance of this approximation in detecting out of distribution examples.  

Let us denote by $z^N$ the training samples, where each $z$ contains both $x$ - a data feature, and $y$ - a label we would like to predict. Also, we will denote by $\Theta$ the class of hypotheses. We will also assume that there is some loss function, $l(y,x,\theta)$ that measures the accuracy of the prediction.

Denote by $\theta^*$ the empirical risk minimizer (ERM) - e.g, the hypothesis that achieves minimal loss over the training set $z^N$. Following \cite{koh2017influence}, adding a new example $z_{add}$ with an infinitesimally small weight $\epsilon$ to the training set will result in the following ERM:

\begin{equation}
    \theta_{new} = \theta^* - \epsilon H_{\theta^*}^{-1} \nabla l(y_{add}, x_{add}, \theta^*).
\end{equation}

where $H_{\theta^*}$ is the hessian matrix of the average loss over $z^N$, and the gradient is that of the loss additional example $z_{add}$. The influence over the loss on another exmaple, $z_{test}$, can also be derived:

\begin{equation}
    l(z_{test}, \theta_{new}) = l(z_{test}, \theta^*) - \epsilon \nabla l(z_{test}, \theta^*)^{T}H_{\theta^*}^{-1} \nabla l(z_{add}, \theta^*).
\end{equation}

Seeing as we will use the logarithmic loss function, $l(y,x,\theta) = -log(p_{\theta}(y|x))$, this translates to:

\begin{equation}
    \label{IF_probability}
    p(y|x) = p_{\theta}(y|x) e^{\epsilon \nabla l(z_{test}, \theta^*)^{T}H_{\theta^*}^{-1} \nabla l(z_{add}, \theta^*)}
\end{equation}

To use this in the context of the pNML approach, for each given test feature $x$, we will use \ref{IF_probability} for each possible $y$, and then normalize to get a valid probability assignment:

\begin{equation}
\label{IF_pNML}
    q(y|x, z^N) = \frac{p_{\theta}(y|x)e^{\epsilon \nabla l(y, x, \theta^*)^{T}H_{\theta^*}^{-1} \nabla l(y, x, \theta)}}{\sum_{\tilde{y}}p_{\theta}(\tilde{y}|x) e^{\epsilon \nabla l(\tilde{y}, x, \theta^*)^{T}H_{\theta^*}^{-1} \nabla l(\tilde{y}, x, \theta)}}
\end{equation}

Throughout this paper, We will refer to the probability assignment in \ref{IF_pNML} either as an approximation to the pNML using the influence function or as a Newton step. Note that indeed, \ref{IF_probability} can be interpreted as conduction Newton step in the (opposite) direction of the gradient of the loss for each possible label, then normalizing to get a valid probability assignment.

The computational cost of this method can be divided to two parts: The first one consists of evaluating the hessian over the training set, $H$, and inverting it. While this may be a cumbersome computation, it only has to happen once, right after the training, and we do not need to repeat it for every test. The other part consists of computing the gradients $\nabla(y,x,\theta^*)$ for each possible $y$ and for each test data feature $x$. This part does have to happen for each different data feature $x$, but it only consists of gradient computation for each possible $y$, which is a small computational price compared to the algorithm presented in $\cite{bibas2019new}$ where several epochs with both the training set and the new test set were performed for each possible $y$. 

Compared to conducting a simple gradient descent step over the new test data feature for each possible label, both methods compute the gradients of the loss function for each possible outcome $y$. However, to compute \ref{IF_pNML} one has to compute the $\nabla l(z_{test}, \theta^*)^{T}H_{\theta^*}^{-1} \nabla l(z_{add}, \theta^*)$ for each $y$, whereas after conducting a simple gradient step one will have to run the updated network for each of the possible values of $y$. 

\section{Theoretical Results}
\label{sec:Exampels}
In this section we will derive the influence-function version of the pNML for various hypotheses classes. We first consider the case of linear regression, for which an exact solution can be found. We then consider both logistic regression and neural networks for which only the last layer is perturbed.

\subsection{Linear Regression}

Consider the case of linear regression, where $p_{\theta}(y|x) = \frac{e^{\frac{-(y-\theta^T x)^2}{2\sigma^2}}}{\sqrt{2*\pi*\sigma^2}}$. Note that the logarithmic loss yields the squared distance between $y$ and $\theta^T x$, up to a multiplicative and additive factors independent of $x$, $y$ and $\theta$. Thus, if we denote by $\Vec{y} = [y1, ..., y^N]$ the labels vector and by $X = [x_1, ..., x_N]$ the input matrix, we get that the hessian matrix and the gradient are as follows:

\begin{equation}
    \label{linear_regression_hessian}
    H_{Z^N} = \frac{X^T X}{2\sigma^2}
\end{equation}
\begin{equation}
    \label{linear_regression_gradient}
    \nabla l(\tilde{y}, x, \theta) = \frac{-x^T(y-\theta^T x)}{2\sigma^2}
\end{equation}

Plugging in \ref{IF_probability}
\begin{equation}
    \label{new_probability function}
    p(y|x) = \frac{1}{\sqrt{2\pi \sigma^2}}e^{\frac{-(y-\theta^T x)(1 - \epsilon x^T (X^T X)^{-1} x)(y-\theta^T x)}{2\sigma^2}}
\end{equation}

This directly leads to

\begin{equation}
    q(y|x) = \frac{e^{\frac{-(y-\theta^T x)^2}{2\tilde{\sigma}^2}}}{\sqrt{2*\pi*\tilde{\sigma^2}}}
\end{equation}

where:
\begin{equation}
    \tilde{\sigma} = \frac{\sigma}{\sqrt{1 - \epsilon x^T (X^T X)^{-1}x}}.
\end{equation}

Note that this result is actually very similar to the  pNML for linear regression presented in \cite{bibas2019new}. The main difference is that here we got the $\epsilon$ factor because... another difference is that here the matrix $X$ does not include the new example $x$, whereas in \cite{bibas2019new} $X =[x_1, ..., x_N, x]$.

\subsection{Logistic Regression}

Consider a binary logistic regression model where $y \in \{0, 1\}$ and $p_{\theta}(y=1|x) = \frac{e^{\theta^T x}}{1+ e^{\theta^T x}}$. In this case, the logarithmic loss function yields:

\begin{equation}
    l(y,x,\theta) = - y \log(\frac{e^{\theta^T x}}{1+ e^{\theta^T x}}) -(1-y) \log({1}{1+ e^{\theta^T x}}) = -y \theta^T x + \log (1+e^{\theta^T x}).
\end{equation}

Thus, the gradient of the loss with respect to $\theta$ is:
\begin{equation}
    \nabla l(y, x, \theta) = x^T(-y + \frac{e^{\theta^T x}}{1+ e^{\theta^T x}}) = x^T(1-p_{\theta}(y|x))
\end{equation}
    And the hessian of the average loss over the training set is:
\begin{equation}
    H[i, j] = \sum_{t=1}^N x_t[i]x_t[j]\frac{e^{\theta^T x}}{(1+ e^{\theta^T x})^2} = \sum_{t=1}^N x_t[i]x_t[j]p_\theta(y|x)(1-p_\theta(y|x))
\end{equation}

\subsection{Single Layer Neural Network with Binary Output}\label{sec:example:binary}

Let us now consider a neural network with a single layer and binary output $y \in \{0, 1\}$. Given some input $x$, the probability that the network will assign to $y=1$ is:
\begin{equation}\label{eq:sigmoid}
    p_{\theta}(y=1|x) = \frac{e^{\sigma(\theta^T x)}}{1 + e^{\sigma(\theta^T x)}}
\end{equation}

where $\sigma(z)$ is some activation function. This yields:

\begin{equation}\label{eq:grad_binary}
    \nabla l(y,x,\theta) = x^T  \dot \sigma(\theta^T x) (-y + \frac{e^{\sigma(\theta^T x)}}{1 + e^{\sigma(\theta^T x)}})
\end{equation}

and the hessian of each sample is:

\begin{equation}\label{eq:hessian_binary}
    H[i,j] = x[i] x[j] \ddot \sigma(\theta^T x)(-y + p_{\theta}(y|x)) + x[i] x[j] \dot \sigma^2(\theta^T x)\frac{e^{ \sigma(\theta^T x)}}{(1 +  e^{\sigma(\theta^T x)})^2}
\end{equation}

Note that this calculation also holds in the case where there is a neural network with several layers, and we want to change only the last layer (e.g, 'freeze' all of the other layers, just as in \cite{zhang2016understanding}. In this case, the $x$'s are not the original data features, but the input of the last layer.

\subsection{Single Layer Neural Network with Multiple Outputs} \label{sec:example:multiple}
Let us now consider a neural network with a single layer and output $y \in \{0,1,...,K\}$. Given some input $x$, the probability that the network will assign to $y=j$ is:
\begin{equation}\label{eq:softmax}
    p_{\theta}(y=j|x) = \frac{e^{\sigma(\theta_j^T x)}}{\sum_{j'=0}^k e^{\theta_{j'}^T x}}
\end{equation}

where $\theta = [\theta_0, ..., \theta_k]$. In this case:

\begin{equation}
    l(y,x,\theta) = -\log(p_\theta(y|x)) =  -\theta_y^T x + \log(\sum_{j'=0}^k e^{\theta_{j'}^T x})   
\end{equation}

Thus, the components of the gradient with respect to a single example are as follows:
\begin{equation}\label{eq:grad_multi}
    \frac{\partial l(t,x,\theta)}{\partial \theta_j[i]} = -\mathbf{1}(y=j) x[i] + \frac{x[i]e^{\theta_j^T x}}{\sum_{j'=0}^k e^{\theta_{j'}^T x}} = -x[i] (\mathbf{1}(y=j) - p_{\theta}(j|x))
\end{equation}

The components of the hessian with respect to a single examples are as follows:

\begin{equation}\label{eq:hessian_multi_1}
    \frac{\partial^2 l(t,x,\theta)}{\partial \theta_j[i_1] \theta_j[i_2]} = \frac{x[i_1]x[i_2]e^{\theta_j^T x}}{\sum_{j'=0}^k e^{\theta_{j'}^T x}} -  \frac{x[i_1]x[i_2]e^{2\theta_j^T x}}{(\sum_{j'=0}^k e^{\theta_{j'}^T x})^2} = x[i_1]x[i_2]p_{\theta}(j|x)(1-p_{\theta}(j|x)) 
\end{equation}

\begin{equation}\label{eq:hessian_multi_2}
    \frac{\partial^2 l(t,x,\theta)}{\partial \theta_j[i_1] \theta_l[i_2]} = -\frac{x[i_1]x[i_2]e^{\theta_j^T x}e^{\theta_l^T x}}{(\sum_{j'=0}^k e^{\theta_{j'}^T x})^2} = -x[i_1]x[i_2]p_{\theta}(j|x)p_{\theta}(l|x) 
\end{equation}

\section{Experimental Results} \label{sec:experiments}
This section describes the applications of the the influence functions in deep neural networks. In order to evaluate our method, we conduct multiple experiments and compare our method, i.e. performing a single Newton step over the last layer of a pre-trained models, versus a simple gradient descent step using the same samples. We also compare our results to those obtained by the simply taking   

We examine our results on the MNIST dataset (\cite{lecun-mnisthandwrittendigit-2010}) and the Fashion-MNIST dataset (\cite{xiao2017fashionmnist}). For all experiments we use the same neural network architecture as shown in Table~\ref{tab:arc}, except for changes to the last layer's size and activation functions which are due to the number of classes.
Our model is comprised of seven layers, not counting the input, where the ReLU activation function (\cite{nair2010rectified}) is applied to the output of every convolutional and fully-connected layer (not including the last layer). Our input is a single channel image (gray scale) of size 32x32. The first layer is a 2-dimensional convolutional layer (labeled as CONV2D) with 32 filters of size 3x3 with a stride of 1. The second layer is the same 2-dimensional convolutional layer with 64 filters. The next layer is a max-pooling layer with a kernel of size 2x2, followed by a dropout layer (\cite{gal2016dropout}) with probability of 0.25. The  The next layer is a standard flatten layer that converts the 64 feature maps to a one dimensional layer of size 9216, followed by a fully-connected layer of size 128. Finally, for the multi-class problem, we use an output layer of size 10 using the softmax (\cite{1641:softmax}) activation function and a categorical cross-entropy loss (\cite{Campbell_onthe}), while for the binary classification problem we use a single neuron with the sigmoid activation function as our last layer, using binary cross-entropy loss.

We build and run our networks using the {\em Keras} (\cite{chollet2015keras}) library with {\em Tensorflow} (\cite{tensorflow2015-whitepaper}) as its backend. We run our network for 12 epochs and train our network using stochastic gradient descent optimizer with a fixed learning rate of 0.01. We use the same learning rate also for the gradient step. In both experiments we train our network using the MNIST training set.

\begin{table}[h]
\scriptsize
  \caption{The architecture of our neural networks.}
  \label{tab:arc}
  \centering
  \begin{tabular}{| c | c |}
    \hline
    \textbf{Layer Type} & \textbf{Input/Output Shape}\\ \hline
     Input & 32, 32, 1\\ \hline
     Conv2D & 26, 26, 32\\ \hline
     Conv2D & 24, 24, 64\\ \hline
     MaxPool2D & 12, 12, 64\\ \hline
     Dropout & 12, 12, 64\\ \hline
     Flatten & 9216\\ \hline
     Fully Connected & 128\\ \hline
     Softmax/Sigmoid & 10/1\\ \hline
\end{tabular}
\end{table}

\subsection{Multi-class Classification}
We train our neural network over the 10 classes of the MNIST training set and achieve an accuracy of 98.62\%. Then, we use the last layer of our pre-trained neural network, which is a single fully connected layer with input of size 128, output of size 10 with the softmax activation function, to evaluate our method. We generate 2 types of evaluation sets; the first one contains 1000 samples of the MNIST test set, the second one contains 1000 samples of the fashion-MNIST test set and is considered as the out-of-distribution sets. For each sample we apply over the last layer both a single gradient descent step and a single Newton step, based the hessian and gradient equations derived in section ~\ref{sec:example:multiple}. For each sample, we compute both steps 10 times, each time using a different label.

For the gradient step for each sample and each possible label, we first compute the gradient by applying equation~\ref{eq:grad_multi}, then we compute the new wights by simply subtract the multiplication of the gradients with the learning rate from the original weights, and finally we compute the unnormalized predictions (i.e. the output of the softmax layer) by applying equation~\ref{eq:softmax} using the new weights. For evaluation purposes, we compute the sum of unnormalized predictions over the possible labels for each sample. Finally, for each sample we compute the new probability by simply divide each label's prediction by the some of all labels predictions (equation~\ref{pNML}).

For the Newton step, we first compute the Hessian of each sample of the training set by using equations~\ref{eq:hessian_multi_1}, \ref{eq:hessian_multi_2}, then we compute the average Hessian of the training sets by summarize all samples Hessians and dividing it by the number of the training set samples. Then for each sample of the evaluation set and each label we calculate new predictions by applying equation~\ref{IF_probability} (denoted as unnormalized predictions). For evaluation purposes, we compute the sum of unnormalized predictions over the possible labels for each sample. Finally, for each sample we compute the new probability (or pNML) by using equation~\ref{IF_pNML}.

Since the hessian of the problem is not necessarily invertible, we have chosen to add to it a small damping term $\lambda I$, in accordance with \cite{koh2017influence}. We have used $\lambda = 0.0001$ for our experiments.

Note that \ref{IF_pNML} contains a parameter $\epsilon$ we have to choose, and that naturally it is clear that the approximation is wrong if at least one of the probabilities given by $\ref{IF_probability}$ is larger than $1$. Thus, we have chosen $\epsilon$ by computing the maximal $\epsilon$ for which on of the probabilities is $1$, and multiplying it by 0.5. 

Table~\ref{tab:results_multi} summarize our results. For each type of step and each evaluation set, we compute the average and standard deviation of both the sum of unnormalized predictions probabilities of the samples, and the maximum of the new predictions probabilities (pNML in the case of Newton step) of the samples.

\begin{table}[h]
\scriptsize
  \caption{Summarize of our results for the multi-class problem.}
  \label{tab:results_multi}
  \centering
  \begin{tabular}{| c | c | c | c |}
    \hline
    \textbf{} & \textbf{Original Predictions} & \textbf{Gradient Step} & \textbf{Newton Step}\\ \hline \hline
    \multicolumn{4}{|c|}{\textbf{In-distribution Test Set (MNIST)}}\\\hline 
     Avg. Sum of unnormalized probabilities & - & 1.012 & 1.042\\ \hline
     Std of Sum of unnormalized probabilities & - & 0.120 & 0.080\\ \hline
     Avg. Max of probabilities & 0.987& 0.984& 0.957\\ \hline
     Std of Max of probabilities & 0.054& 0.073& 0.083\\ \hline \hline
     \multicolumn{4}{|c|}{\textbf{Out-of-distribution Test Set (Fashion-MNIST)}}\\\hline 
     Avg. Sum of unnormalized probabilities & - & 1.354 & 1.518  \\ \hline
     Std of Sum of unnormalized probabilities & - & 0.502 &  0.308 \\ \hline
     Avg. Max of probabilities & 0.741& 0.677& 0.579\\ \hline
     Std of Max of probabilities & 0.211& 0.254& 0.193\\ \hline

\end{tabular}
\end{table}

\begin{figure}[H]
\begin{subfigure}{.5\textwidth}
  \centering
  \includegraphics[width=.9\linewidth]{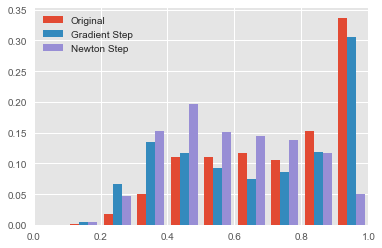}
  \caption{Max Probabilities}
\end{subfigure}%
\begin{subfigure}{.5\textwidth}
  \centering
  \includegraphics[width=.9\linewidth]{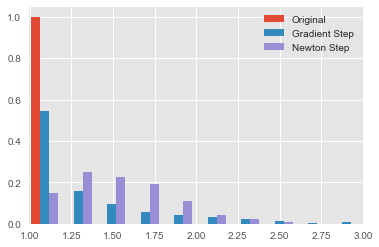}
  \caption{Sum of Unnormalized Probabilities}
\end{subfigure}
\caption{Histograms of (a) Max Probabilities and (b) Sum of Unnormalized Probabilities over the Out of Distribution Set, for; original predictions, gradient step and Newton step.}
\label{fig:histograms_out_multi}
\end{figure}

As table~\ref{tab:results_multi} shows there is only a small difference between both the gradient step, Newton step and the original predictions when dealing with the test set. However, over the out-of-distribution there are significant differences between the difference methods as can be seen in Fig.~\ref{fig:histograms_out_multi}.

\begin{figure}[H]
\begin{subfigure}{.5\textwidth}
  \centering
  \includegraphics[width=.9\linewidth]{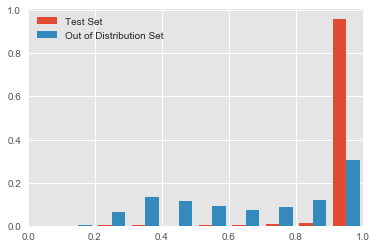}
  \caption{Gradient Step}
\end{subfigure}%
\begin{subfigure}{.5\textwidth}
  \centering
  \includegraphics[width=.9\linewidth]{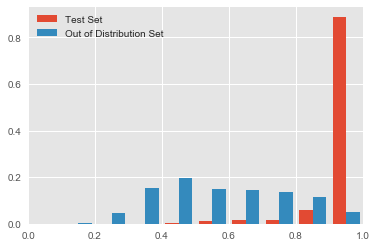}
  \caption{Newton Step}
\end{subfigure}
\caption{Histograms of Max Probabilities using (a) gradient step and (b) Newton step for; test set and out-of-distribution set.}
\label{fig:histograms_comparison_multi}
\end{figure}


Furthermore, it can be seen (Fig.~\ref{fig:histograms_comparison_multi}) that using the Newton step we achieve a significant difference between the distribution of the in-distribution set and the out-of-distribution set, far more significant than the gradient step.

\subsection{Binary Classification}
We train our neural network over the following 2 classes of the MNIST training set; 'six' and 'nine', and achieve an accuracy of 99.75\%. Then, we use the last layer of our pre-trained neural network, which is a single neuron with input of size 128, output of size 1 with the Sigmoid activation function, to evaluate our method. We generate 3 types of evaluation sets; the first one contains 1000 samples of the 'six' and 'nine' classes of the MNIST test set, the second one contains 974 samples of the 'eight' class of the MNIST test set and is considered as the additional class set (or in-distribution set), and the third contains 1000 samples of the fashion-MNIST test set and is considered as the out-of-distribution set. For each sample we apply over the last layer both a single gradient descent step and a single Newton step, based on Sec.~\ref{sec:example:binary}. For each sample, we compute both steps twice, such that each time we compute using different label.

For the gradient step, for each sample and each possible label, we first compute the gradient by applying equation~\ref{eq:grad_binary}, then we compute the new wights by simply subtract the multiplication of the gradients with the learning rate from the original weights, and finally we compute the unnormalized prediction (i.e. the output of the Sigmoid) by applying equation~\ref{eq:sigmoid} using the new weights. For evaluation purposes, we compute the sum of the unnormalized predictions over the two labels for each sample. Finally, for each sample we compute the new probability by simply dividing each label's prediction by the some of the two labels predictions (equation~\ref{pNML}).

For the Newton step, we first compute the Hessian of each sample of the training set by using equation~\ref{eq:hessian_binary}, then we compute the average Hessian of the training sets by summarize all samples Hessians and dividing it by the number of the training set samples. Then for each sample of the evaluation set and each label we calculate new predictions by applying equation~\ref{IF_probability} (denoted as unnormalized predictions). For evaluation purposes, we compute the sum of unnormalized predictions over the possible labels for each sample. Finally, for each sample we compute the new probability (or pNML) by using equation~\ref{IF_pNML}.

Table~\ref{tab:results_binary} summarize our results. For each method and each evaluation set, we compute the average and standard deviation of both the sum of unnormalized predictions probabilities of the samples, and the maximum of the new predictions probabilities of the samples.

\begin{table}[h]
\scriptsize
  \caption{Summarize of our results for the binary classification problem.}
  \label{tab:results_binary}
  \centering
  \begin{tabular}{| c | c | c | c |}
    \hline
    \textbf{} & \textbf{Original Predictions} & \textbf{Gradient Step} & \textbf{Newton Step}\\ \hline \hline
    \multicolumn{4}{|c|}{\textbf{In-distribution Test Set (MNIST: Classes 'six' and 'nine')}}\\\hline 
     Avg. Sum of unnormalized probabilities & - & 1.005 & 1.018\\ \hline
     Std of Sum of unnormalized probabilities & - & 0.028 & 0.043\\ \hline
     Avg. Max of probabilities & 0.995 & 0.992& 0.981\\ \hline
     Std of Max of probabilities & 0.034& 0.042& 0.051\\ \hline \hline
     \multicolumn{4}{|c|}{\textbf{In-distribution Additional Class Set (MNIST: class 'eight')}}\\\hline 
     Avg. Sum of unnormalized probabilities & - & 1.099 & 1.122 \\ \hline
     Std of Sum of unnormalized probabilities & - & 0.143 &  0.119 \\ \hline
     Avg. Max of probabilities & 0.931& 0.881& 0.861\\ \hline
     Std of Max of probabilities & 0.119& 0.155& 0.146\\ \hline \hline
     \multicolumn{4}{|c|}{\textbf{Out-of-distribution Test Set (Fashion-MNIST)}}\\\hline 
     Avg. Sum of unnormalized probabilities & - & 1.233 & 1.171 \\ \hline
     Std of Sum of unnormalized probabilities & - & 0.234 &  0.114 \\ \hline
     Avg. Max of probabilities & 0.899& 0.779& 0.803\\ \hline
     Std of Max of probabilities & 0.132& 0.172& 0.145\\ \hline

\end{tabular}
\end{table}

\begin{figure}[H]
\begin{subfigure}{.5\textwidth}
  \centering
  \includegraphics[width=.9\linewidth]{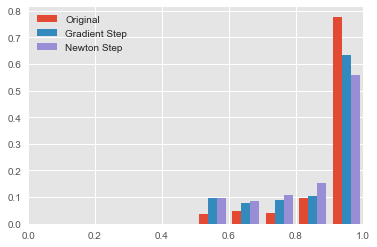}
  \caption{Max Probabilities}
\end{subfigure}%
\begin{subfigure}{.5\textwidth}
  \centering
  \includegraphics[width=.9\linewidth]{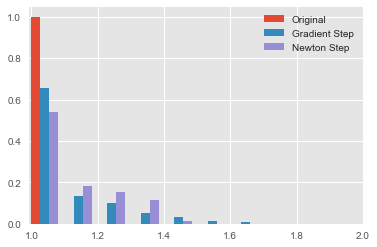}
  \caption{Sum of Unnormalized Probabilities}
\end{subfigure}
\caption{Histograms of (a) Max Probabilities and (b) Sum of Unnormalized Probabilities over the Additional Class Set, for; original predictions, gradient step and Newton step.}
\label{fig:histograms_add_binary}
\end{figure}

\begin{figure}[H]
\begin{subfigure}{.5\textwidth}
  \centering
  \includegraphics[width=.9\linewidth]{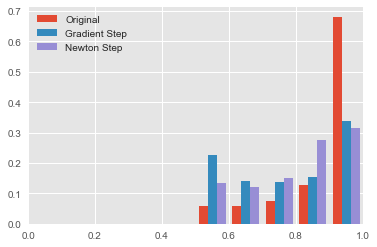}
  \caption{Max Probabilities}
\end{subfigure}%
\begin{subfigure}{.5\textwidth}
  \centering
  \includegraphics[width=.9\linewidth]{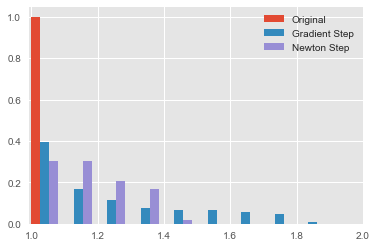}
  \caption{Sum of Unnormalized Probabilities}
\end{subfigure}
\caption{Histograms of (a) Max Probabilities and (b) Sum of Unnormalized Probabilities over the Out of Distribution Set, for; original predictions, gradient step and Newton step.}
\label{fig:histograms_out_binary}
\end{figure}

As table~\ref{tab:results_binary} shows there is only a small difference between both the gradient step, Newton step and the original predictions when dealing with the test set. However, over the additional class set and the out-of-distribution there are significant differences between the difference methods as can be seen in Fig.~\ref{fig:histograms_add_binary} and ~\ref{fig:histograms_out_binary}.

\begin{figure}[H]
\begin{subfigure}{.5\textwidth}
  \centering
  \includegraphics[width=.9\linewidth]{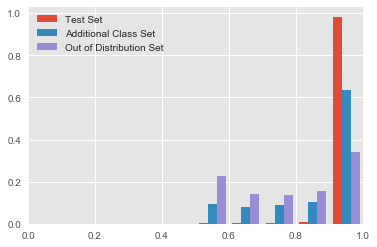}
  \caption{Gradient Step}
\end{subfigure}%
\begin{subfigure}{.5\textwidth}
  \centering
  \includegraphics[width=.9\linewidth]{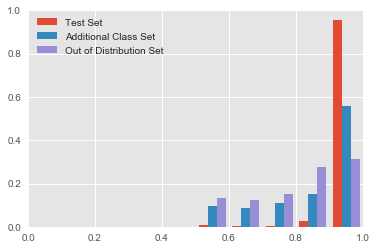}
  \caption{Newton Step}
\end{subfigure}
\caption{Histograms of Max Probabilities using (a) gradient step and (b) Newton step for; test set, additional class set and out-of-distribution set.}
\label{fig:histograms_comparison_binary}
\end{figure}


Furthermore, it can be seen (Fig.~\ref{fig:histograms_comparison_binary}) that using both the Newton step and the gradient step we achieve a significant difference between the distribution of the in-distribution sets and the out-of-distribution set. Moreover, by looking at the histogram, it's easy to distinguish between the different classes. 

\section{Concluding Remarks} \label{sec:conclusions}
In this work we proposed and analyzed an approximation of the pNML which is based on influence functions. we derived the influence-function version of the pNML for various hypotheses classes including those of a single layer neural network with binary output and a single layer neural network with binary multiple outputs. By conducting multiple experiments and performing a single Newton step over the last layer of a pre-trained binary and multi-class models, we showed that when applied to neural networks, this approximation can detect out-of-distribution examples effectively. Specifically, we showed that by analyzing the statistics of the unnormalized and normalized probabilities, we can distinguish between in-distribution and out-of-distribution sets. Furthermore, we compared the proposed Newton step with a simple gradient descant step, and showed that the first yield much significant difference between the two sets. 

As for directions for future work, it would be interesting to compare the performance of our method, both in terms of computation time and in terms of detection accuracy, to the performance of other methods for out-of-distribution detection. It can also be interesting to further investigate how the choice of parameters $\lambda$ and $\epsilon$ effects the detection performance.

\FloatBarrier
\bibliographystyle{apalike} 
\bibliography{ms.bbl}

\appendix
\end{document}